\documentclass[journal,twoside,web]{ieeecolor}
\usepackage{generic}
\usepackage{cite}
\usepackage{multirow}
\usepackage{amsmath,amssymb,amsfonts}
\usepackage{algorithmic}
\usepackage{graphicx}
\usepackage{algorithm,algorithmic}
\usepackage{hyperref}
\hypersetup{hidelinks=true}
\usepackage{textcomp}
\usepackage{hyperref}
\hypersetup{hypertex=true,
            colorlinks=true,
            linkcolor=blue,
            anchorcolor=blue,
            citecolor=blue}

\usepackage{amsmath,amsfonts}
\usepackage{algorithmic}
\usepackage{algorithm}
\usepackage{array}
\usepackage[caption=false,font=small,labelfont=rm,textfont=rm]{subfig}
\usepackage{textcomp}
\usepackage{stfloats}
\usepackage{float}
\usepackage{url}
\usepackage{verbatim}
\usepackage{graphicx}
\usepackage{flushend}
\usepackage{cite}
\usepackage{booktabs}
\usepackage{makecell}
\usepackage{cite}
\usepackage{amsmath,amssymb,amsfonts}
\usepackage{algorithmic}
\usepackage{graphicx}
\usepackage{textcomp}
\usepackage{cleveref}
\def\BibTeX{{\rm B\kern-.05em{\sc i\kern-.025em b}\kern-.08em
    T\kern-.1667em\lower.7ex\hbox{E}\kern-.125emX}}
\begin{document}
\title{A Semi-Supervised Framework for Breast Ultrasound Segmentation with Training-Free Pseudo-Label Generation and Label Refinement}
\author{{Ruili Li, \IEEEmembership{Student Member, IEEE}, Jiayi Ding\textsuperscript{*}, Ruiyu Li, Yilun Jin\textsuperscript{*}, Shiwen Ge, Yuwen Zeng, \IEEEmembership{Member, IEEE}, Xiaoyong Zhang, \IEEEmembership{Senior member, IEEE}, Eichi Takaya, Jan Vrba and Noriyasu Homma, \IEEEmembership{Member, IEEE}}
\thanks{This work was supported in part by JSPS KAKENHI Grant Numbers JP18K19892, JP20K08012,
JP19H04479 and JSTSPRING, GrantNumber JPMJSP2114.}
\thanks{Ruili Li, Jiayi Ding, Eichi Takaya and Noriyasu Homma are with the Tohoku University Graduate School of Medicine, Sendai 980-8575, Japan (e-mail: li.ruili.t3@dc.tohoku.ac.jp, ding.jiayi.s2@dc.tohoku.ac.jp, 
homma@ieee.org).} 
\thanks{Yuwen Zeng is with Advanced Institute of Convergence Knowledge Informatics, Research Institute of Electrical Communication, Tohoku University, Sendai 980-0812, Japan (e-mail: yuwen@tohoku.ac.jp)}
\thanks{Xiaoyong Zhang is with the National Institute of Technology, Sendai College, Sendai 989-3128, Japan (e-mail: xiaoyong@ieee.org).}
\thanks{Ruiyu Li is with State Key Laboratory of Oncology in South China, Sun Yat-sen University Cancer Center, Guangzhou 510060, China (e-mail: liry2@sysucc.org.cn).}
\thanks{Jan Vrba is with the Department of Mathematics, Informatics,
and Cybernetics, University of Chemistry and Technology, Prague,
Czech Republic (e-mail:
Jan.Vrba@vscht.cz).}
\thanks{Shiwen Ge is with  School of Software Technology, Zhejiang University,Ningbo 315100,China (e-mail: 22451129@zju.edu.cn)}
\thanks{Yilun Jin is with Southeast University, School of Cyber Science and Engineering Nanjing, Jiangsu, China (email: jinyilun@seu.edu.cn)}
\thanks{This work has been submitted to the IEEE for possible publication. Copyright may be transferred without notice, after which this version may no longer be accessible.}
\thanks{\textsuperscript{*}Corresponding author: ding.jiayi.s2@dc.tohoku.ac.jp, jinyilun@seu.edu.cn}
}
\maketitle

\begin{abstract}
Semi-supervised learning (SSL) has emerged as a promising paradigm for breast ultrasound (BUS) image segmentation, but it often suffers from unstable pseudo labels under extremely limited annotations, leading to inaccurate supervision and degraded performance. Recent vision–language models (VLMs) provide a new opportunity for pseudo-label generation, yet their effectiveness on BUS images remains limited because domain-specific prompts are difficult to transfer. 

To address this issue, we propose a semi-supervised framework with training-free pseudo-label generation and label refinement. By leveraging simple appearance-based descriptions (e.g., ``dark oval''), our method enables cross-domain structural transfer between natural and medical images, allowing VLMs to generate structurally consistent pseudo labels. These pseudo labels are used to warm up a static teacher that captures global structural priors of breast lesions. Combined with an exponential moving average teacher, we further introduce uncertainty–entropy weighted fusion and adaptive uncertainty-guided reverse contrastive learning to improve boundary discrimination. 

Experiments on four BUS datasets demonstrate that our method achieves performance comparable to fully supervised models even with only 2.5\% labeled data, significantly outperforming existing SSL approaches. Moreover, the proposed paradigm is readily extensible: for other imaging modalities or diseases, only a global appearance description is required to obtain reliable pseudo supervision, enabling scalable semi-supervised medical image segmentation under limited annotations.

\end{abstract}


\section{Introduction}
\label{sec:intro}

Breast ultrasound (BUS) imaging \cite{huang2017breast} is one of the most widely used non-invasive imaging modalities for the detection and diagnosis of breast lesions \cite{guo2018ultrasound}. Owing to its real-time imaging capability, absence of ionizing radiation, and cost-effectiveness, BUS has become a routine tool in clinical practice for breast cancer screening and diagnosis \cite{sood2019ultrasound,cheng2010automated}. Therefore, the precise segmentation of breast cancer lesions in BUS images plays a crucial role in assisting early diagnosis. To achieve this, various automated segmentation approaches have been developed, among which fully supervised deep learning methods have recently achieved remarkable success \cite{zhang2023fully,xue2021global}. However, these methods rely on large amounts of high-quality pixel-wise labels, which are extremely time-consuming to obtain and require expert radiologists. To alleviate this annotation burden, semi-supervised learning (SSL) has been developed \cite{zhai2022ass,pdfnet}.

Despite the progress of existing SSL methods, they still face notable limitations on BUS images. First, many SSL frameworks rely on pseudo labels, making training vulnerable to early prediction errors and confirmation bias \cite{su2024mutual}. Second, under few labeled samples, the teacher model is under-trained and produces noisy pseudo labels, offering weak supervision to the student \cite{pan2024dual}. Third, consistency regularization often relies on strong–weak augmentations designed for RGB natural images \cite{yang2023revisiting,luo2021semi}, which are less suitable for grayscale, speckle-noisy BUS data. These problems become more severe in the extremely low-label regime. Methods such as mean teacher (MT) or confidence-filtered pseudo labeling often produce fragmented predictions with inaccurate boundaries. In addition, confidence thresholds may select “high-confidence mistakes,” which prevents pseudo labels from forming a stable lesion prior and limits SSL performance under scarce annotations.

\begin{figure}[t]
  \centering
  \includegraphics[width=\columnwidth]{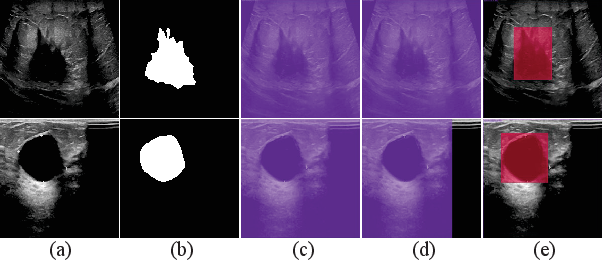}
  \caption{To illustrate how different textual prompts affect zero-shot knowledge transfer, we visualize bounding box generated from (c) medical terms (``tumor''), (d) radiological attributes (``high density''), and (e) appearance-based descriptions(``dark oval.dark round.dark lobulated'').
}
  \label{fig:introbbx}
\end{figure}

To provide a stronger pseudo-label prior in the extremely low-label regime, it is natural to seek an external model that can directly produce pseudo masks instead of relying solely on the under-trained teacher. Recent vision–language and vision foundation models (e.g., Grounding DINO\cite{liu2024grounding} and SAM \cite{kirillov2023segment}) can generate text-guided boxes and masks on natural images, enabling automatic pseudo-label generation. However, transferring them to BUS remains challenging in two aspects \cite{segmedi,du2025medical}.
(i) Directed segmentation via training-based adaptation. Fine-tuning these large models (or DINO-style adaptations) to directly perform BUS segmentation is misaligned with our target setting: it typically requires (1) bounding-box or region-level annotations to bootstrap training, (2) a relatively large amount of labeled data for stable fine-tuning, and (3) instance-specific vision–text pairs or customized descriptions \cite{segmedi,du2025medical}. These requirements are unrealistic under clinical scenarios. (ii) Pseudo-label generation via training-free transfer. Direct zero-shot prompting is also unreliable. Medical and radiological prompts often fail because general VLMs lack domain semantics, and BUS images are grayscale with speckle noise and weak boundaries, leading to unstable and structurally inconsistent pseudo masks.
Furthermore, in SSL-based BUS segmentation, prediction errors and domain mismatch are often concentrated in hard regions (e.g., weak or ambiguous lesion boundaries), yielding large low-confidence areas with high uncertainty and noisy pseudo masks. However, many recent semi-supervised segmentation methods \cite{jiang2024ph,wang2022U2PL} apply contrastive learning by sampling global or “reliable” pixel features, which tends to neglect these uncertain yet informative regions, limiting robustness improvements. Despite these challenges, we observe that BUS lesions still exhibit relatively consistent appearance traits that can serve as useful cues for prompt design. Leveraging such traits, appearance-based prompts enable more reliable training-free pseudo-label generation, while subsequent label refinement can further mitigate the noise and uncertainty in these pseudo masks (Fig.~\ref{fig:introbbx}).

Inspired by this insight, we propose a semi-supervised framework for BUS segmentation with training-free pseudo-label generation and label refinement. Specifically, appearance-based prompts are used to guide vision–language models to generate structurally meaningful pseudo labels without additional training. These pseudo labels provide an external structural prior that alleviates the instability of pseudo-label generation from under-trained teachers in extremely low-label regimes. 

To further improve pseudo-label quality, we introduce a label refinement strategy within a dual-teacher semi-supervised framework, where predictions from a VLM-initialized teacher and an EMA teacher are adaptively fused based on uncertainty and entropy. In addition, an uncertainty-guided reverse contrastive learning mechanism is employed to focus on hard regions and enhance boundary discrimination.

Our main contributions are summarized as follows:
\begin{itemize}

\item 
We propose a training-free pseudo-label generation strategy for BUS segmentation that leverages appearance-based prompts to guide VLMs in producing structurally meaningful pseudo labels, enabling effective cross-domain structural transfer from natural images.

\item 
We develop a semi-supervised framework with a label refinement mechanism that integrates a VLM-initialized teacher and an EMA teacher through uncertainty–entropy weighted fusion, together with adaptive uncertainty-guided reverse contrastive learning to improve pseudo-label reliability and boundary discrimination.

\item 
Extensive experiments on  public BUS datasets demonstrate that our method consistently outperforms existing semi-supervised approaches and achieves performance comparable to fully supervised models. Notably, with only 2.5\% labeled data, our method approaches the fully supervised upper bound, showing strong robustness in the extreme low-label regime.

\end{itemize}
\section{Related Work}
\subsection{Semi-Supervised Image Segmentation}
Semi-supervised segmentation reduces annotation cost by learning from unlabeled images. Many effective methods have been proposed. MT \cite{tarvainen2017MT} establishes a teacher–student consistency framework and serves as a widely used baseline. U2PL \cite{wang2022U2PL} improves pseudo-label quality via entropy-based confidence estimation. BCP \cite{bai2023BCP} narrows distribution gaps through bidirectional copy–paste mixing between labeled and unlabeled data. MCF \cite{wang2023mcf} stabilizes training with multi-level feature consistency, while CSC-PA \cite{ding2025csc} enhances semantic coherence through cross-sample prototype alignment. For BUS segmentation specifically, PH-Net \cite{jiang2024ph} emphasizes hard-region learning to address fuzzy boundaries, PGCL \cite{basak2023pseudo} incorporates pseudo-label-guided contrastive learning to mine informative patches, and AAU \cite{adiga2024anatomically} introduces anatomically-aware uncertainty estimation to guide semi-supervised learning.

Despite their success, most semi-supervised methods still derive supervision from the model itself (e.g., pseudo labels or confidence maps), making them vulnerable to early errors, especially for BUS images with fuzzy boundaries and speckle noise. As a result, obtaining more reliable pseudo labels remains a key challenge for semi-supervised BUS segmentation.
\subsection{VLM-Guided and Foundation Segmentation Models}
Beyond conventional SSL, recent studies leverage foundation models and VLM priors to reduce annotation cost via text-guided localization and promptable segmentation. UniSeg \cite{butoi2023universeg} leverages  large-scale pretrained representations for universal segmentation across datasets. In SSL, CLIP-style vision--text alignment has also been used to enhance pseudo-label reliability and feature discrimination \cite{huang2025text}. Meanwhile, prompt-based adaptation of SAM has also been explored for medical image segmentation, such as SAM-MediCLIPV2 \cite{koleilat2025medclip}, which aligns SAM with medical domain knowledge via vision--language pretraining to improve prompt-to-mask transfer in clinical images. However, these foundation or VLM-based pipelines are mostly developed and tuned on breast datasets via domain-aligned pretraining or adaptation, which typically demands large-scale breast ultrasound data to reliably align the model. This requirement becomes particularly impractical in the extremely low-label regime.
\section{Methodology}
\subsection{Overview}
\begin{figure*}[htbp]
  \centering
  \includegraphics[width=\textwidth]{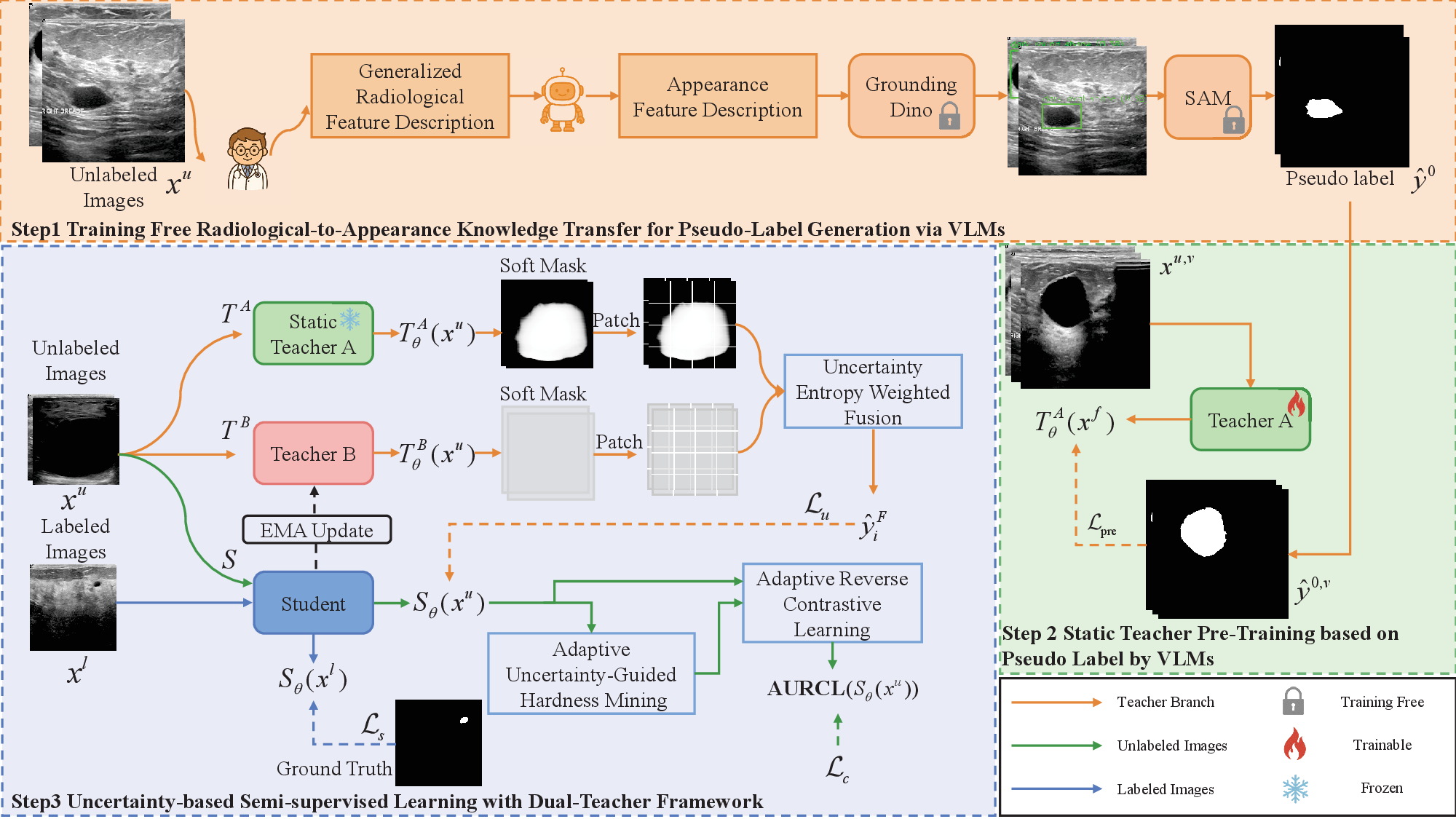}
  \caption{Overview of the proposed semi-supervised BUS segmentation framework.
The pipeline consists of two stages: (1) Appearance-Prompted Pseudo-Label Generation (APPG), where appearance-prompted vision–language models (VLMs) produce initial pseudo labels in a training-free manner; 
(2) Pseudo-Label Refinement, which includes two steps: static-teacher warm-up training to capture coarse structural priors of breast lesions, and uncertainty-based semi-supervised learning using a dual-teacher framework with Uncertainty–Entropy Weighted Fusion (UEWF) and Adaptive Uncertainty-Guided Reverse Contrastive Learning (AURCL).}
  \label{fig:model}
\end{figure*}

Our semi-supervised segmentation framework is trained using a small labeled dataset $x^l$ and a large unlabeled dataset $x^u$. The overall architecture is illustrated in Fig.~\ref{fig:model}. The method consists of two key components: training-free pseudo-label generation and pseudo-label refinement. First, to provide reliable supervision under extremely limited annotations, we generate pseudo labels using vision–language models guided by appearance-based prompts. These pseudo labels provide coarse yet structurally meaningful lesion priors and are used to pre-train a static teacher model $T^A$, which remains frozen during subsequent training. Second, to refine these pseudo labels and improve segmentation quality, we adopt a dual-teacher semi-supervised framework composed of the frozen static teacher $T^A$, a dynamic teacher $T^B$, and a student model $S$. The dynamic teacher $T^B$ is updated via exponential moving average (EMA) \cite{ema} of the student parameters to maintain temporal consistency. During training, the student model is supervised by ground-truth annotations for labeled data $x^l$, while for unlabeled data $x^u$, the teachers provide pseudo labels to guide learning. To stabilize early training, the student model is initialized with the pretrained weights of $T^A$. The supervised loss $\mathcal{L}_s$ on $x^l$ is computed using a combination of Binary Cross-Entropy (BCE) and Dice losses, while the unsupervised loss $\mathcal{L}_u$ on $x^u$ adopts the same formulation using teacher-generated pseudo labels. In addition, a contrastive loss $\mathcal{L}_c$ is introduced to enhance feature discrimination, particularly in ambiguous boundary regions. The overall objective is defined as
\begin{equation}
\mathcal{L} = \mathcal{L}_s + \lambda_u \mathcal{L}_u + \lambda_c \mathcal{L}_c ,
\end{equation}
where $\lambda_u$ and $\lambda_c$ control the contributions of the unsupervised and contrastive losses, respectively.
\subsection{Step 1: Appearance-Prompted Training-Free Pseudo-Label Generation (APPG) via VLMs}
In breast tumor segmentation, obtaining sufficient pixel-level annotations is extremely challenging, which makes it difficult for semi-supervised models to learn reliable representations. 
As a result, the pseudo labels generated by teacher–student frameworks often suffer from low quality and structural inconsistency, which significantly degrades the effectiveness of the student model during training. 
Moreover, most existing semi-supervised segmentation methods based on consistency regularization or strong–weak data augmentation are originally designed for natural RGB images \cite{zhao2023augmentation}. 
Such augmentations and perturbations do not transfer well to grayscale medical images like BUS images, since variations of lesion boundaries are domain-specific. 
Consequently, common teacher–student frameworks exhibit poor generalization and unstable pseudo labels when directly applied to breast tumor segmentation tasks.

In recent years, vision–language models (VLMs) have demonstrated strong capabilities in text-guided object localization and segmentation on natural images, making them promising tools for automatic pseudo-label generation.

However, directly transferring these models to medical image analysis remains challenging. 
Most VLMs are primarily trained on large-scale RGB natural image datasets with object-centric semantics, and therefore lack exposure to grayscale medical data and domain-specific terminology. 
As a result, when medical terms or anatomical descriptions are used as prompts, these models often fail to interpret the semantics correctly and struggle to localize lesion regions.

Unlike many other medical imaging modalities, BUS images exhibit relatively consistent appearance patterns. 
Lesions typically present clear contrast against surrounding tissues and maintain regular geometric shapes, such as oval or round regions. 
These visual characteristics can be effectively represented through simple appearance-based textual descriptions that capture the overall shape and brightness patterns of lesions. Such appearance-based prompts provide intuitive structural guidance for VLMs, enabling effective cross-domain transfer and facilitating training-free pseudo-label generation for BUS images.

To implement the proposed appearance-prompted training-free pseudo-label generation (APPG), we first summarize common medical and radiological characteristics of breast tumors rather than case-specific features.
These general traits include typical medical term, lesion shapes, boundary clarity, and overall brightness contrast, such as ``high density'', ``tumor'', ``heterogeneous hypoechoic texture'' and ``spiculated margins'', which are consistently observed in clinical ultrasound interpretation. 
A large language model (LLM) \cite{achiam2023gpt} is then employed to transform these domain-level features into concise and interpretable natural-language expressions. 
Instead of generating individual captions for each image, the LLM (GPT-5 in our implementation) produces a compact set of universal appearance descriptions, such as ``dark oval,'' ``dark round,'' and ``dark lobulated,'' which capture the shared visual appearance patterns of breast tumors. 
These descriptions are embedded into the prompts of a  VLMs to perform training-free segmentation on unlabeled BUS images with training free manner.

Formally, given an unlabeled BUS image $x_i^u$, the appearance-based generalized prompt $aprmpt$ is defined as:
\begin{equation}
aprmpt = \text{LLM}(med\_common\_traits)
\label{prmpt}
\end{equation}
where $\text{med\_common\_traits}$ denotes the set of domain-level imaging descriptors describing tumor shape, boundary, and brightness. 
The generated prompt $aprmpt$ serves as textual guidance for the VLMs to produce a coarse pseudo mask $\hat{y}_i^{0}$ in a training-free manner:
\begin{equation}
\hat{y}_i^{0} = \text{VLM}(x_i^u, aprmpt)
\label{vlm}
\end{equation}

Here, Eq.~(\ref{prmpt}) represents the conversion from domain knowledge to general appearance descriptions via the LLM, while Eq.~(\ref{vlm}) denotes the generation of an initial pseudo label through training-free inference of the VLMs guided by those descriptions. 
The resulting pseudo masks $\hat{y}_i^{0}$ provide coarse but structurally meaningful lesion priors for subsequent refinement and static-teacher pre-training.

Specifically, in our implementation, the generated appearance descriptions and corresponding BUS images are embedded into the VLMs implemented using Grounding DINO. 
The VLMs performs text-guided object localization to produce bounding boxes $b_i$ that roughly indicate potential lesion regions according to the appearance descriptions:
\begin{equation}
b_i^u = \mathrm{VLM}_{\mathrm{DINO}}(x_i^u, aprmpt)
\end{equation}
Each predicted bounding box $b_i^u$, together with the raw BUS image $x_i^u$, is then fed into the SAM to generate the corresponding segmentation mask:
\begin{equation}
\hat{y}_i^{0} = \text{SAM}(x_i^u, b_i^u)
\end{equation}
Through this two-stage process, APPG leverages the localization capability of Grounding DINO and the segmentation ability of SAM to produce VLMs-based pseudo labels in a fully training-free manner.

\subsection{Pseudo-Label Refinement}
\subsubsection{Step 2: Refinement via Static Teacher Warm-up Pre-training}
Although the APPG process provides coarse pseudo masks in a training-free manner, 
some generated masks may be missing or incomplete due to the limited transferability of VLMs to ultrasound images. 
To provide a reliable prior for the subsequent refinement stage, a filtering strategy is first applied to remove invalid pseudo labels before the static-teacher warm-up training.

Formally, let $x^u = \{x_i^u\}_{i=1}^{N}$ denote the unlabeled dataset and $\hat{y}_i^{0}$ the corresponding pseudo masks generated by APPG. 
We define the valid subset $x^u_{\text{valid}}$ as:
\begin{equation}
x^u_{\text{valid}} = \{(x_i^u, \hat{y}_i^{0}) \mid \text{area}(\hat{y}_i^{0}) > \tau \},
\end{equation}
where $\text{area}(\hat{y}_i^{0})$ denotes the proportion of foreground pixels in the pseudo mask. 
The threshold $\tau$ is empirically set to $1\%$ of the image area to remove empty or invalid masks. 
Only the remaining samples with structurally valid pseudo labels are retained for the warm-up training stage.

Based on the filtered subset $x^u_{\text{valid}}$, the static teacher $T^A$ is initialized and trained using the VLM-generated pseudo labels to capture coarse structural priors of lesion regions. 
For each valid sample $(x_i^{u}, \hat{y}_i^{0}) \in x^u_{\text{valid}}$, the static teacher predicts the segmentation map and is optimized using a combination of Binary Cross-Entropy (BCE) and Dice losses. 
Here $T_{\theta}^{A}$ denotes the static teacher network parameterized by $\theta$. 

After this warm-up stage, the parameters of $T^A$ are frozen and used as the static teacher in the subsequent pseudo-label refinement process within the semi-supervised framework.

\subsubsection{Step 3: Refinement via Uncertainty-Based Semi-Supervised Learning}
\paragraph{Uncertainty entropy weighted fusion}
In our framework, pseudo labels  originate from two complementary sources:  
(1) the static teacher $T^A$, initialized through warm-up training with pseudo labels generated by  APPG, and  
(2) the dynamic teacher $T^B$, which evolves by receiving exponential moving average (EMA) updates from the student model $S$.
Considering the inherent difference between these two types of supervision — the former being structurally reliable but less adaptive, and the latter being temporally consistent but potentially noisy, we design a dual-teacher mechanism to jointly exploit their strengths.

At each iteration, both teachers generate soft pseudo labels for the same unlabeled image $x_i^{u}$, 
denoted as $\hat{\mathbf{y}}_i^{A}$ and $\hat{\mathbf{y}}_i^{B}$, respectively. 
The student model $S$ then learns from both labeled data and these dual sources of soft pseudo labels. 
However, since the static and dynamic teachers exhibit different levels of reliability across region, an Uncertainty--Entropy Weighted Fusion (UEWF) strategy is introduced to handle the inconsistency between pseudo labels generated by two different teachers.

Given the soft pseudo labels $\hat{\mathbf{y}}_i^{A}$ and $\hat{\mathbf{y}}_i^{B}$ generated by the static teacher $T^A$ and the dynamic teacher $T^B$,
we quantify the pixel-wise predictive uncertainty by the Shannon entropy\cite{shannon1948mathematical} of the class posterior:
\begin{equation}
\mathcal{H}\!\left(\hat{\mathbf{y}}(\mathbf{p})\right)
= - \sum_{c=1}^{C} \hat{\mathbf{y}}_{c}(\mathbf{p})\,\log\!\left(\hat{\mathbf{y}}_{c}(\mathbf{p})+\epsilon\right),
\label{eq:entropy}
\end{equation}
where $\epsilon$ is a small constant for numerical stability. $\mathbf{p}$ denotes the predicted pixel-wise probability obtained after the sigmoid activation, $c \in \{1,\dots,C\}$ denotes the class index (with $C$ classes).
The corresponding uncertainty map $\mathbf{E}_A$ and $\mathbf{E}_B$ are smoothed via patch-wise average pooling ($k=14$) to suppress local noise and emphasize region-level reliability:
\begin{equation}
\mathbf{E}_{A,B}^{\text{smooth}} = \text{Upsample}\big(\text{AvgPool}(\mathbf{E}_{A,B}, k)\big)
\tag{8}
\end{equation}
Confidence weights are then computed as the inverse of the smoothed entropy:
\begin{equation}
\mathbf{w}_{A,B} = \frac{1}{\mathbf{E}_{A,B}^{\text{smooth}} + \epsilon}
\tag{9}
\end{equation}
Finally, the fused pseudo label is obtained as the uncertainty–aware weighted average:
\begin{equation}
\hat{\mathbf{y}}_i^{F} =
\frac{\mathbf{w_A} \cdot \hat{\mathbf{y}}_i^{A} + \mathbf{w_B}  \cdot \hat{\mathbf{y}}_i^{B}}
{\mathbf{w_A} + \mathbf{w_B} + \epsilon}
\tag{10}
\end{equation}
The fused labels $\hat{\mathbf{y}}_i^{F}$ are used as reliable supervisory signals for the student model during the unsupervised training stage.
\paragraph{Adaptive uncertainty-guided reverse contrastive learning}
Recent semi-supervised segmentation studies \cite{wang2022U2PL, jiang2024ph} have applied contrastive learning to enhance feature discrimination by sampling global or reliable pixel features.
However, these methods often neglect hard or uncertain regions, which are crucial for improving model robustness.
Although some recent work \cite{jiang2024ph}  attempts to utilize unreliable pixels, its strategy fails to address uncertainty adaptively.
In contrast, our adaptive uncertainty–guided reverse contrastive learning (AURCL) module explicitly focuses on the uncertain yet informative low-confidence regions predicted by the student model.
By reversely augmenting and contrasting these pixels, AURCL encourages the network to refine ambiguous boundaries and learn more discriminative representations for challenging areas.

\begin{figure}[htbp]
  \centering
  \includegraphics[width=\columnwidth]{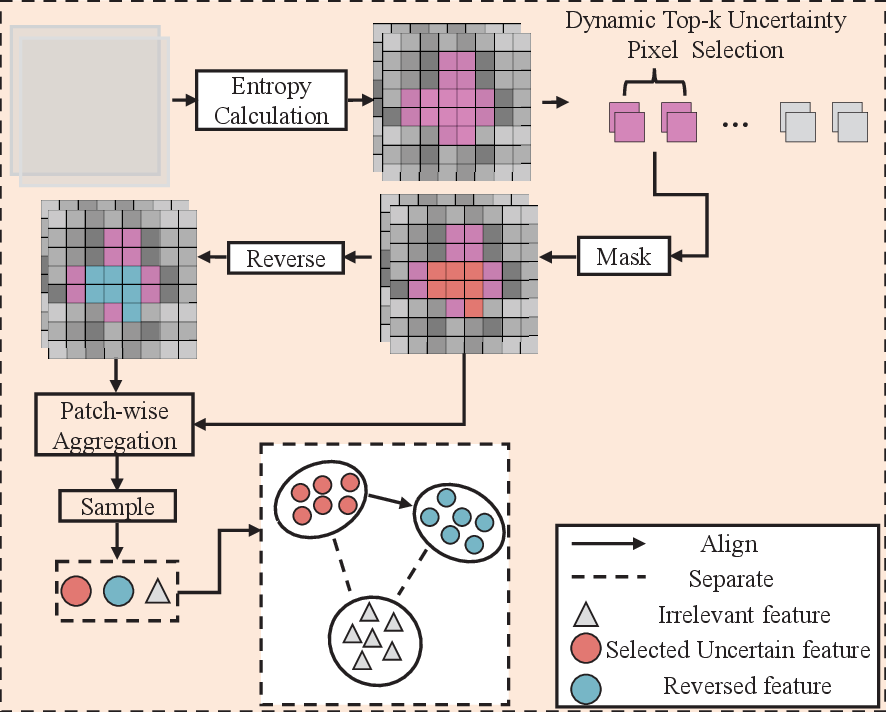}
  \caption{Structure of the proposed AURCL module. Entropy maps identify top-$k$ high-uncertainty pixels, whose predictions are reversed and aggregated into patch-level features. Contrastive learning then pulls the original and reversed high-entropy features from these high-entropy patches closer to each other, while pushing them apart from stable low-entropy patches.}
  \label{fig:ARUCL}
\end{figure}

Fig.\ref{fig:ARUCL} shows our proposed AURCL module. For an unlabeled image $x_i^u$, the predicted probability map is given by 
$\mathbf{p}_i = \sigma(S_\theta(x_i^u))$, where $\sigma(\cdot)$ denotes the sigmoid function. We denote by $\mathbf{F}_i$ the feature map extracted from the student network that produces the probability map $\mathbf{p}_i$.
The confidence map is defined as $\mathbf{C}_i = 1 - \mathbf{p}_i$. 

To accurately locate uncertain pixels, we introduce a dynamic top-K threshold $\tau_i$ that combines a percentile-based adaptive threshold and a fixed lower bound ($\tau_{\text{fix}} = 0.2$): 
\begin{equation}
\tau_i = \max\!\Big(\operatorname{top-K}(\mathbf{C}_i, K),\,\tau_{\text{fix}}\Big),
\qquad K= rHW
\tag{11}
\label{eq:tau_threshold}
\end{equation}
where $r$ denotes the reverse ratio controlling the proportion of low-confidence pixels.
This adaptive threshold ensures consistent selection of uncertain pixels across samples
while preventing overconfidence during early training, thereby maintaining sufficient
low-confidence pixels for effective contrastive learning.

Then we create low-confidence mask $\mathbf{M}_i^{\text{low}}$, and perform a probability reversal within these pixels, and get the uncertain pixel-wise probability map  $\tilde{\mathbf{p}_i}$:
\begin{equation}
\mathbf{M}_i^{\text{low}}(u,v)=\mathbb{I}\!\left(\mathbf{C}_i(u,v)\ge \tau_i\right),
\tag{12}
\end{equation}

\begin{equation}
\begin{aligned}
\tilde{\mathbf{p}}_i(u,v) &= \big(1-\mathbf{M}_i^{\text{low}}(u,v)\big)\odot\mathbf{p}_i(u,v) \\
&\quad + \mathbf{M}_i^{\text{low}}(u,v)\odot\big(1-\mathbf{p}_i(u,v)\big)
\end{aligned}
\tag{13}
\end{equation}

where $(u,v)$ denotes the pixel location, $\mathbf{p}_i(u,v)\in[0,1]$ is the predicted foreground probability, and $\tilde{\mathbf{p}}_i(u,v)$ is the reversed probability at the selected low-confidence pixels. $\mathbb{I}(\cdot)$ denotes the indicator function, returning 1 if the condition holds and 0 otherwise (and $\odot$ denotes element-wise multiplication when written in matrix form).

Such reversal operation enables the network to distinguish between well-learned and poorly-learned pixels in the feature space, laying the foundation for the subsequent contrastive learning refinement in AURCL. After constructing a reversed view for high-uncertainty regions, we extract patch-level features ${\mathbf f}_{i,j}$ and ${\tilde{\mathbf f}}_{i,j}$  via average pooling to aggregate local evidence and mitigate pixel-level speckle noise. 
\begin{equation}
\begin{aligned}
{\mathbf f}_{i,j} &=
\frac{\sum_{(u,v)\in\Omega_j}\mathbf{F}_i(u,v)\,\mathbf{p}_i(u,v)}
{\sum_{(u,v)\in\Omega_j}\mathbf{p}_i(u,v)+\epsilon}, \\
\tilde{{\mathbf f}}_{i,j} &=
\frac{\sum_{(u,v)\in\Omega_j}\mathbf{F}_i(u,v)\,\hat{\mathbf{p}}_i(u,v)}
{\sum_{(u,v)\in\Omega_j}\hat{\mathbf{p}}_i(u,v)+\epsilon}.
\end{aligned}
\tag{14}
\end{equation}

where $\Omega_j$ denotes the pixel set of the $j$-th patch, where $\mathbf{F}_i(u,v)$ denotes the feature vector at pixel location $(u,v)$ on the feature map  $\mathbf{F}_i$

The module is then optimized with a contrastive objective that aligns features from the same spatial patch across the original and reversed views, while pushing apart features from mismatched patches. This enforces cross-view consistency and encourages more stable representations around ambiguous boundaries. The contrastive loss is defined as:

\begin{equation}
\mathcal{L}_{\text{AURCL}}
=
-\frac{1}{N}\sum_{j=1}^{N}
\log
\frac{\exp\!\left(\mathrm{sim}(\mathbf{f}_{i,j},\tilde{\mathbf{f}}_{i,j})/\tau\right)}
{\sum\limits_{k=1}^{N}
\exp\!\left(\mathrm{sim}(\mathbf{f}_{i,j},\tilde{\mathbf{f}}_{i,k})/\tau\right)},
\label{eq:aurcl_infonce}
\end{equation}
where $N$ is the number of patches in one image, $\mathbf f_{i,j}$ and $\tilde{\mathbf f}_{i,j}$ denote the patch-level features of the $j$-th patch in the original and reversed views of the $i$-th sample, respectively. For each anchor patch, $(\mathbf f_{i,j},\tilde{\mathbf f}_{i,j})$ forms the positive pair from the same spatial location, while $\{\tilde{\mathbf f}_{i,k}\}_{k\neq j}$ from other patch locations in the same image act as negatives. $\tau$ is the temperature, and $\mathrm{sim}(\mathbf a,\mathbf b)=\frac{\mathbf a^\top \mathbf b}{\|\mathbf a\|\|\mathbf b\|}$ denotes cosine similarity.

\section{Experiments}
\label{sec:experiments}
\subsection{Datasets}

\noindent\textbf{BUSI}. The BUSI dataset \cite{busi} contains 780 breast ultrasound images collected from Baheya Hospital in Cairo, Egypt, consisting of 133 normal, 437 benign, and 210 malignant cases.  Following prior works, we use only the 647 abnormal images (benign and malignant) for segmentation experiments.

\noindent\textbf{UBB (UDIAT + BREASTUSG + BUSUCLM).}
To evaluate cross-dataset generalization, we merge three BUS datasets into a combined set termed UBB.
UDIAT \cite{udiat} includes 163 breast ultrasound images from the UDIAT Diagnostic Center, containing 110 benign and 53 malignant lesions.
BREASTUSG \cite{breastusg} consists of 256 breast ultrasound scans collected from 256 patients and 266 benign and malignant segmented lesions. 
BUSUCLM \cite{Vallez2024BUSUCLMBU} comprises 38 breast cancer patients, encompassing a total of 174 benign and 90 malignant images. Both datasets are randomly divided into training, validation, and test subsets at an 8:1:1 ratio. The training sets are further partitioned into labeled and unlabeled subsets with labeled ratios of 2.5\%, 10\%, and 20\%. For the fully unsupervised baseline that trains U-Net using only VLMs-generated pseudo labels, we  only retrain images for valid masks. The effective sample counts are 499 for BUSI and 474 for UBB, since images without VLM outputs (no detected boxes or masks) are excluded.
\subsection{Evaluation Metrics}
We adopt three widely used segmentation metrics to evaluate model performance: 
the Dice, IoU, and Acc. 

\subsection{Implementation Details.}
All experiments were conducted on a single NVIDIA GeForce RTX 4090 GPU using the PyTorch framework with a fixed random seed. Our network and all compared models adopt ResNet-34 \cite{he2016deep}. The input image size is 224$\times$224 and training is performed for 100 epochs with a batch size of 8, where labeled and unlabeled samples are equally included in each batch. The Adam optimizer is used with a momentum of 0.995, and the learning rate is adjusted by a ReduceLROnPlateau scheduler. No data augmentation is applied to our model, while augmentation-based methods follow the weak–strong strategies described in their original papers. The hyperparameters $\lambda_{u}$ and $\lambda_{c}$ are set to 1 and 0.5, respectively.
\subsection{Comparison with State-of-the-Art methods}

We compare our method with six state-of-the-art semi-supervised segmentation approaches, including MT \cite{tarvainen2017MT}, U2PL \cite{wang2022U2PL}, BCP \cite{bai2023BCP}, PH-Net \cite{jiang2024ph}, MCF \cite{wang2023mcf}, and CSC-PA \cite{ding2025csc},PGCL\cite{basak2023pseudo}, Text-semiseg\cite{huang2025text} and AaU-ssm\cite{adiga2024anatomically}. 
To ensure fairness, all methods are implemented under the same experimental protocol and identical data preprocessing and augmentation strategies, and all hyperparameters are tuned according to the optimal settings reported in their original papers or open-source implementations.

As shown in Table~\ref{tab:busi_ubb_results}, our proposed method achieved the highest scores on all three evaluation metrics on two datasets, with labeled data settings of 2.5\%, 10\%, and 20\%, respectively. On the BUSI dataset, our method outperformed the previous state-of-the-art method across all label proportions, achieving Dice improvements of +13.79\% (2.5\% labeled), +2.34\% (10\% labeled), and +2.55\% (20\% labeled). On the more challenging UBB dataset, which involves multi-source and cross-device ultrasound imaging, the performance gains are even more substantial, with Dice improvements of +15.99\%, +1.25\%, and +2.6\% under the 2.5\%, 10\%, and 20\% label settings, respectively.
\begin{figure*}[!htbp]
  \centering
  \includegraphics[width=\textwidth]{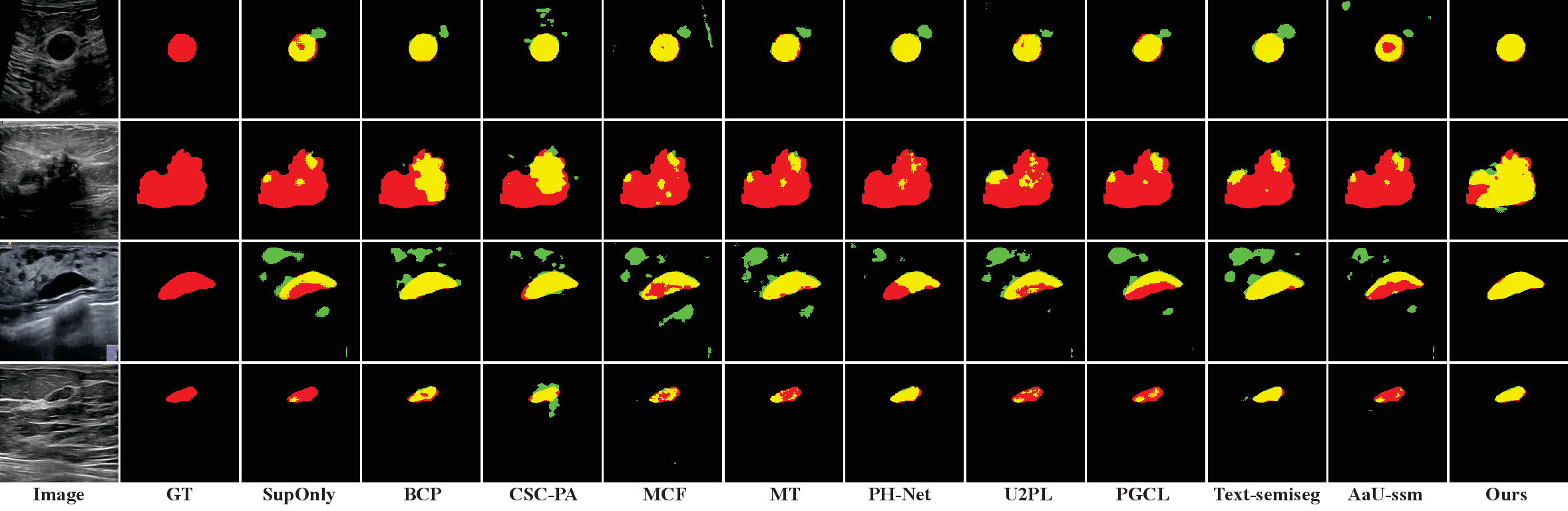}
  \caption{Visual comparison of different state-of-the-art methods on the BUSI and UBB datasets. All models are trained in 2.5\% labeled data. Red, green and yellow regions represent ground truth, prediction and overlapping regions, respectively.}
  \label{fig:duibitu}
\end{figure*}
\begin{table*}[t]
  \centering
  \caption{Comparison of segmentation performance on BUSI and UBB (UDIAT + BREASTUSG + BUSUCLM) datasets under different labeling ratios.}
  \label{tab:busi_ubb_results}
  \resizebox{0.94\textwidth}{!}{
  \begin{tabular}{llcccccccccc}
    \toprule
    \multirow{2}{*}{Method} & \multirow{2}{*}{Venue} &
    \multicolumn{5}{c}{BUSI} &
    \multicolumn{5}{c}{UBB (UDIAT + BREASTUSG + BUSUCLM)} \\
    \cmidrule(lr){3-7} \cmidrule(lr){8-12}
     &  & Labeled & Unlabeled & Dice(\%)$\uparrow$ & IoU(\%)$\uparrow$ & Acc(\%)$\uparrow$ &
       Labeled & Unlabeled & Dice(\%)$\uparrow$ & IoU(\%)$\uparrow$ & Acc(\%)$\uparrow$ \\
    \midrule

    U-Net (static teacher) & -- & 0 & 499 & 67.34 & 57.28 & 93.03 & 0 & 474 & 69.95 & 60.26 & 96.67 \\
    \midrule

    U-Net & MICCAI & 12 (2.5\%) & 0 & 50.00 & 39.71 & 92.35 & 13 (2.5\%) & 0 & 43.52 & 31.03 & 93.52 \\
    U-Net & MICCAI  & 51 (10\%) & 0 & 65.04 & 54.77 & 94.76 & 54 (10\%) & 0 & 67.39 & 55.76 & 96.75 \\
    U-Net & MICCAI  & 103 (20\%) & 0 & 70.94 & 61.79 & 95.06 & 108 (20\%) & 0 & 67.95 & 58.22 & 96.93 \\
    U-Net & MICCAI  & 517 (100\%) & 0 & \textbf{81.68} & \textbf{73.74} & \textbf{96.65} &
                          542 (100\%) & 0 & 74.81 & 65.56 & \textbf{97.29} \\
    \bottomrule

    MT\cite{tarvainen2017MT} & NeurIPS'17 & & & 54.09 & 42.93 & 92.87 & & & 56.73 & 43.43 & 94.36 \\
    U2PL\cite{wang2022U2PL} & CVPR'22 & & & 51.53 & 41.12 & 93.26 & & & 48.57 & 35.38 & 93.98 \\
    BCP\cite{bai2023BCP} & CVPR'23 & & & 58.93 & 49.48 & 93.89 & & & 56.95 & 44.72 & 93.10 \\
    MCF\cite{wang2023mcf} & CVPR'23  & & & 49.18 & 39.30 & 92.33 & & & 49.29 & 36.26 & 94.20 \\
    PH-Net\cite{jiang2024ph} & CVPR'24  & 12 (2.5\%) & 505 (97.5\%) & 55.13 & 45.28 & 92.79 &
                                13 (2.5\%) & 529 (97.5\%) & 50.02 & 38.82 & 94.66 \\
    CSC-PA\cite{ding2025csc} & CVPR'25  & & & 58.78 & 45.97 & 93.68 & & & 51.06 & 37.28 & 93.71 \\
    PGCL\cite{basak2023pseudo} & CVPR'23  & & & 54.26 & 43.31 & 92.88 & & & 43.73 & 31.39 & 93.69 \\
    Text-semiseg\cite{huang2025text} & MICCAI'25  & & & 56.85 & 45.35 & 93.13 & & & 59.76 & 46.30 & 93.97 \\
    AaU-ssm\cite{adiga2024anatomically} & MedIA'24  & & & 53.75 & 42.82 & 92.83 & & & 45.51 & 32.76 & 94.37 \\
    Ours & -- & & & \textbf{72.72} & \textbf{63.11} & \textbf{95.08} &
             & & \textbf{75.75} & \textbf{65.86} & \textbf{96.67} \\
    \bottomrule

    MT\cite{tarvainen2017MT} & NeurIPS'17 & & & 68.12 & 58.48 & 94.93 & & & 65.79 & 55.15 & 96.71 \\
    U2PL\cite{wang2022U2PL} & CVPR'22 & & & 67.85 & 57.93 & 94.73 & & & 70.88 & 59.44 & 96.85 \\
    BCP\cite{bai2023BCP} & CVPR'23 & & & 66.65 & 57.22 & 94.98 & & & 72.39 & 60.59 & 96.87 \\
    MCF\cite{wang2023mcf} & CVPR'23  & & & 66.71 & 56.74 & 95.10 & & & 65.61 & 54.23 & 96.44 \\
    PH-Net\cite{jiang2024ph} & CVPR'24  & 51 (10\%) & 466 (90\%) & 72.64 & 63.54 & 95.19 &
                                54 (10\%) & 488 (90\%) & 73.89 & 61.85 & 96.99 \\
    CSC-PA\cite{ding2025csc} & CVPR'25  & & & 66.41 & 55.51 & 94.68 & & & 72.65 & 61.08 & 96.90 \\
    PGCL\cite{basak2023pseudo} & CVPR'23  & & & 63.89 & 53.50 & 94.66 & & & 66.04 & 54.83 & 96.61 \\
    Text-semiseg\cite{huang2025text} & MICCAI'25  & & & 75.06 & 65.88 & 95.46 & & & 74.70 & 63.85 & 96.90 \\
    AaU-ssm\cite{adiga2024anatomically} & MedIA'24  & & & 53.75 & 42.82 & 92.83 & & & 63.51 & 52.10 & 96.21 \\
    Ours & -- & & & \textbf{77.40} & \textbf{68.50} & \textbf{95.71} &
             & & \textbf{75.95} & \textbf{67.09} & \textbf{97.13} \\
    \bottomrule

    MT\cite{tarvainen2017MT} & NeurIPS'17 & & & 73.60 & 64.27 & 95.68 & & & 72.42 & 62.53 & 97.15 \\
    U2PL\cite{wang2022U2PL} & CVPR'22 & & & 71.69 & 62.38 & 95.44 & & & 75.46 & 65.27 & 97.23 \\
    BCP\cite{bai2023BCP} & CVPR'23 & & & 69.48 & 61.03 & 95.42 & & & 71.62 & 61.00 & 96.98 \\
    MCF\cite{wang2023mcf} &  CVPR'23  & & & 73.23 & 63.83 & 95.58 & & & 70.42 & 60.47 & 96.83 \\
    PH-Net\cite{jiang2024ph} &  CVPR'24  & 103 (20\%) & 414 (80\%) & 70.24 & 61.48 & 95.69 &
                                108 (20\%) & 434 (80\%) & 75.12 & 65.04 & 97.20 \\
    CSC-PA\cite{ding2025csc} & CVPR'25  & & & 70.44 & 59.36 & 95.58 & & & 74.18 & 62.47 & 96.95 \\
    PGCL\cite{basak2023pseudo} &  CVPR'23  & & & 71.40 & 62.82 & 95.52 & & & 72.14 & 62.45 & 97.25 \\
    Text-semiseg\cite{huang2025text} & MICCAI'25  & & & 75.83 & 67.25 & 95.92 & & & 75.55 & 65.90 & 97.24 \\
    AaU-ssm\cite{adiga2024anatomically} & MedIA'24  & & & 71.55 & 62.93 & 95.71 & & & 70.86 & 59.94 & 96.80 \\
    Ours & -- & & & \textbf{78.38} & \textbf{69.12} & \textbf{95.93} &
             & & \textbf{78.15} & \textbf{68.82} & \textbf{97.27} \\
    \bottomrule

  \end{tabular}
}
\end{table*}

Importantly, under the extremely limited labeled data setting (2.5\%), our method demonstrates particularly strong capability in leveraging unlabeled data. On the BUSI dataset, our model achieves a Dice of 72.72\%, significantly higher than all other semi-supervised baselines under the same setting. This effect is even more remarkable on the UBB dataset. Using only 13 labeled images, our method achieves a Dice coefficient of 75.75\%, which is 15.99\% higher than the previous best method and even outperforms a fully supervised UNet\cite{ronneberger2015u} trained on 100\% labeled data (74.81\%). This highlights that our appearance-prompted pseudo-labeling and uncertainty-based semi-supervised learning with dual-teacher framework allow the model to learn stable lesion shape and boundary characteristics directly from unlabeled images. Overall, these results indicate that our method  could  achieve accurate and robust lesion segmentation in real clinical settings, especially when high-quality annotations are difficult, costly, or impractical to obtain.

Furthermore, to intuitively demonstrate the superiority of our method, we visualize the segmentation results of several representative breast ultrasound images under the 2.5\% labeled data setting in Fig. \ref{fig:duibitu}. Despite the extremely limited supervision, our approach consistently yields more accurate and coherent delineations than other methods.

\subsection{Ablation studies}
\begin{figure}[t]
  \centering
  \includegraphics[width=\columnwidth]{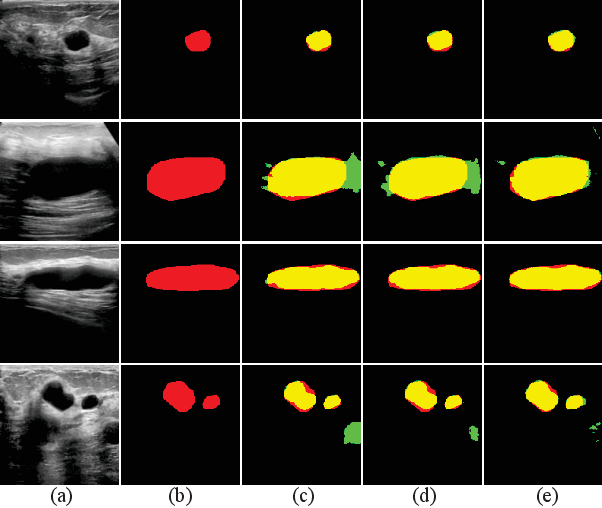}
  \caption{Visual comparison of different ablation study on the BUSI and UBB datasets. All models are trained in 2.5\% labeled
data. Red, green and yellow regions represent ground truth, prediction and overlapping regions, respectively. (a) image, (b) ground truth, (c) APPG (step1), (d) STWP (step 2), (e) Ours}
  \label{fig:abla}
\end{figure}

\noindent \textbf{Effectiveness of components.}
As shown in Table~\ref{tab:method_comparison}, we conduct ablation studies on each component of our proposed framework using the BUSI dataset with 2.5\% labeled data. The supervised baseline trained only on labeled data exhibits low performance (50.00\% Dice, 39.71\% IoU) due to limited supervision. Introducing APPG brings the largest improvement (+14.09\% Dice, +17.57\% IoU), as the  appearance-prompted VLMs pseudo labels provide stable lesion shape and boundary priors and prevent early error accumulation in unsupervised learning stage.
Integrating APPG into the dual-teacher framework results in an additional 3.83\% improvement in Dice and 4.35\% in IoU, indicating that the static teacher offers consistent structural guidance while the dynamic teacher refines local details throughout training. Adding AURCL further improves performance (+0.47\% Dice, +0.86\% IoU) by explicitly enhancing feature discrimination in low-confidence boundary regions.
UEWF provides an additional 0.52\% Dice improvement by fusing teacher predictions based on uncertainty, thereby reducing the influence of unreliable pseudo-label areas. With all components combined, the full model achieves the highest scores in Dice and IoU. Overall, the results clearly verify the importance of each component in achieving robust BUS segmentation under extremely limited annotations.
\begin{table}[htbp]
\centering
\caption{Ablation study of different components on BUSI dataset under 2.5\% partition protocol. “SupOnly” stands for training using only labeled data.}
\resizebox{\linewidth}{!}{
\begin{tabular}{lcccccc}
\toprule
\textbf{Method} & \textbf{APPG}  &\textbf{UEWF} & \textbf{AURCL} & \textbf{Dice(\%)} & \textbf{IoU(\%)} \\
\midrule
SupOnly &  &  & &50.00&39.71 \\
I      &  &\checkmark   & & 54.09 &42.93 \\
III      & \checkmark  & & & 71.17&61.63 \\
IIII     & \checkmark  &  &\checkmark & 71.64&62.49  \\
IIIII    & \checkmark  & \checkmark & & 72.16 & 62.56 \\
IIIIII    & \checkmark  & \checkmark &\checkmark & \textbf{72.72} & \textbf{63.11} \\
\bottomrule
\end{tabular}
}
\label{tab:method_comparison}
\end{table}
Furthermore, to intuitively demonstrate the effectiveness of our proposed components, we visualize representative BUS segmentation results under the 2.5\% labeled setting in Fig.~\ref{fig:abla}. 
As can be seen, training with the initial pseudo labels generated by the appearance-prompted pseudo-label generation (APPG) still yields incomplete and boundary-ambiguous masks due to the extremely limited supervision. 
After incorporating the static-teacher warm-up pre-training (STWP) stage, which learns structural priors from the pseudo labels, the predictions become more coherent but may remain imperfect in highly ambiguous regions. 
By further introducing the uncertainty-based semi-supervised refinement stage (Step 3), including uncertainty–entropy weighted fusion (UEWF) and adaptive uncertainty-guided reverse contrastive learning (AURCL), the final model produces the most accurate and consistent delineations, especially around fuzzy boundaries, consistently outperforming other methods in this ultra-low-label regime.
\begin{table}[t]
\centering
\caption{Ablation study on the effectiveness of the patchify in UEWF under 2.5\% partition protocol.} 
\begin{tabular}{lcc}
\toprule
Method & Dice (\%) & IoU (\%) \\
\midrule
w/o Patchify &71.89  &62.52  \\
w/ Patchify  & \textbf{72.72} & \textbf{63.11} \\
\bottomrule
\end{tabular}
\label{tab:mask_ablation}
\end{table}

\noindent \textbf{Ablation study on hyper-parameters.}
(1) Patchify in UEWF. As shown in Table \ref {tab:mask_ablation}, enabling patch-wise smoothing increases Dice from 71.89\% to 72.72\% and IoU from 62.52\% to 63.11\%. This indicates that region-level uncertainty estimation is more robust to local speckle noise than pixel-wise entropy, leading to more stable fused pseudo labels in BUS images.
(2) Confidence threshold $\tau$ in AURCL. Table \ref{tab:aurcl_thresh} reports the ablation study of the confidence threshold $\tau$ in Eq (\ref{eq:tau_threshold}). As $\tau$ decreases from 60 to 20, Dice shows only slight variation, while IoU consistently increases (from 62.35\% to 63.11\%). This suggests that selecting moderately low-confidence regions enables effective boundary refinement without introducing excessive noise.
\begin{table}[t]
\centering
\caption{Ablation study on the confidence threshold $\tau$ in the AURCL module under 2.5\% partition protocol.}
\begin{tabular}{ccc}
\toprule
$\tau$ & Dice (\%) & IoU (\%) \\
\midrule
60 & 71.81 & 62.35 \\
40 & 71.96 & 62.45 \\
20 & \textbf{72.72} & \textbf{63.11}  \\
\bottomrule
\end{tabular}
\label{tab:aurcl_thresh}
\end{table}

\begin{figure}[!t]
  \centering
  \includegraphics[width=\columnwidth]{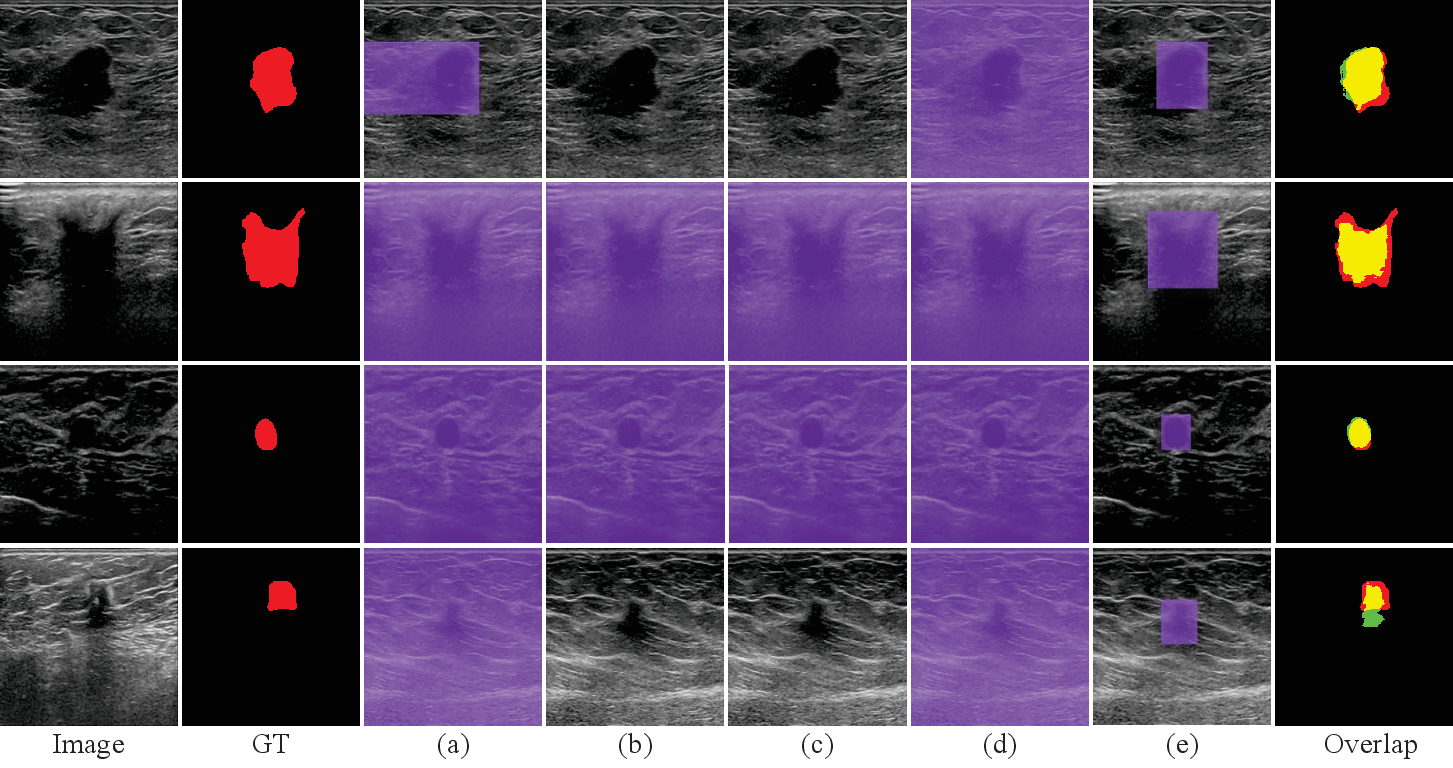}
  \caption{
  Visual comparison of zero-shot bounding boxes and pseudo masks under different prompt conditions. 
  (a) Bounding boxes generated using the medical term \emph{``tumor''}; 
  (b) bounding boxes generated using the medical term \emph{``lesion''}; 
  (c) bounding boxes generated using the medical term \emph{``breast cancer''}; 
  (d) bounding boxes generated using the radiology term \emph{``high density''}; 
  (e) bounding boxes produced by our proposed \emph{``Appearance-prompted training-free pseudo label generation''} strategy; 
  Overlap denotes pseudo masks generated by applying SAM to the bounding boxes from our strategy, and the overlapping regions with GT. 
  }
  \label{fig:zeroshot_combined}
\end{figure}

\noindent \textbf{Visual comparison of zero-shot bounding boxes labeling Under different prompt conditions}
Fig. \ref{fig:zeroshot_combined}   provide a visual comparison of zero-shot bounding boxes generated under different prompt conditions. 
Specifically, subfigures (a)–(d) show the detection results obtained using four commonly used medical or imaging terms, including \emph{``tumor''}, \emph{``lesion''}, \emph{``breast cancer''}, and \emph{``high density''}. 
These prompts exhibit limited performance when applied across datasets or domains, resulting in unstable localization and low accuracy. 
In contrast, (e) illustrates the bounding boxes generated by our proposed appearance-prompted training-free pseudo labeling strategy, which achieves higher precision and consistency across different samples and datasets. 
Furthermore, Overlap presents the pseudo masks generated by applying SAM to the bounding boxes from our strategy, along with the overlapping regions with the ground truth (GT). 
The results clearly demonstrate that our method achieves cross-domain capability, providing strong guidance for localization and generating high-quality pseudo labels.

\noindent \textbf{Comparison with VLM-based Baselines.}
Table~\ref{tab:vlm_fewshot} compares our method with representative VLM-based baselines under the 2.5\% labeled protocol. 
Overall, these pipelines remain clearly limited on BUSI (Dice: 28.74--44.14\%, IoU: 21.99--33.09\%) and UBB (Dice: 23.87--49.67\%, IoU: 18.55--36.40\%). 
One possible reason is that VLM-based approaches tend to generalize well when inter-case appearance variations are relatively small, such as in normal or non-lesion images. 
However, breast tumor regions often exhibit large inter-case variability in shape, size, and boundary patterns, making it difficult for foundation-model priors to consistently localize and segment lesions under extremely limited supervision.

In contrast, our method explicitly introduces stable pseudo-label priors through APPG and further refines them with the proposed semi-supervised framework. 
This substantially strengthens boundary learning and leads to 72.72\% Dice and 63.11\% IoU on BUSI, significantly outperforming the VLM-based baselines.

\begin{table}[t]
\centering
\caption{Comparison with VLM-based few-shot methods on BUSI and UBB under the 2.5\% labeled partition protocol. All methods use 2.5\% ground-truth labels for few-shot setting.}
\label{tab:vlm_fewshot}
\setlength{\tabcolsep}{3.6pt}
\begin{tabular}{l c cc cc}
\toprule
\multirow{2}{*}{\textbf{Method}} & \multirow{2}{*}{\textbf{Venue}} 
& \multicolumn{2}{c}{\textbf{BUSI}} & \multicolumn{2}{c}{\textbf{UBB}} \\
\cmidrule(lr){3-4}\cmidrule(lr){5-6}
& & \textbf{Dice(\%)} & \textbf{IoU(\%)} & \textbf{Dice(\%)} & \textbf{IoU(\%)} \\
\midrule
MediClipV2         & MedIA'25  & 28.74 & 22.08 & 23.87 & 18.55 \\
UniversalSeg       & ICCV'23   & 30.68 & 21.99 & 48.17 & 35.77 \\
GroundingDino(SFT) & ECCV'24   & 44.14 & 33.09 & 49.67 & 36.40 \\
\midrule
\textbf{Ours}      & --        & \textbf{72.72} & \textbf{63.11} & \textbf{75.75} & \textbf{65.86} \\
\bottomrule
\end{tabular}
\end{table}

\begin{figure}[t]
  \centering
  \includegraphics[width=0.8\columnwidth]{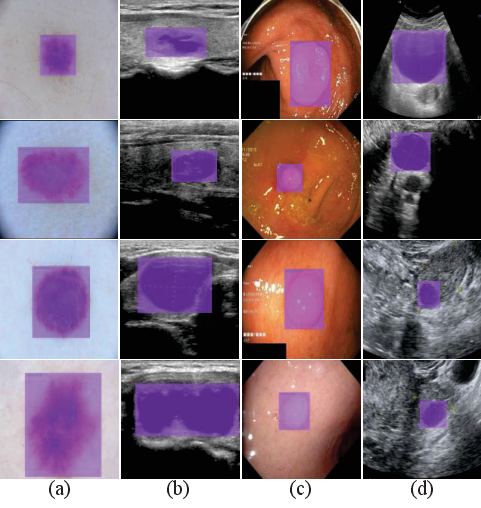}
  \caption{
  Qualitative results on multiple datasets and imaging modalities to demonstrate the generalizability of our APPG pipeline. (a) skin lesion (dermoscopy), (b) thyroid nodule/tumor (ultrasound), (c) colorectal polyp (endoscopy), and (d) ovarian lesion (ultrasound). 
  }
  \label{fig:genral}
\end{figure}

\noindent \textbf{Visual comparison of cross-modal generalization.}Fig.~\ref{fig:genral} shows qualitative results on multiple datasets and imaging modalities, highlighting the robustness of our appearance-prompted training-free pseudo labeling strategy.
For a given disease type, the target region typically exhibits relatively consistent visual patterns (e.g., characteristic shape, boundary sharpness, and different color or lightness), even when collected from different devices or clinical sites.
Therefore, it is often possible to craft a \emph{general} appearance description that remains valid across domains, rather than relying on dataset-specific or diagnosis-driven prompts.
With such generalizable appearance cues, our method produces stable bounding boxes across dermoscopy (skin lesions), ultrasound (thyroid and ovarian lesions), and endoscopy (polyps), which can be directly passed to SAM to generate coherent pseudo masks with clear boundaries. This significantly saves time and labeling cost. When new data or a new domain becomes available, one can quickly bootstrap reliable pseudo supervision using the same (or minimally adjusted) appearance description, reducing expensive pixel-level annotation. More importantly, the resulting pseudo labels can be directly leveraged by our semi-supervised training pipeline, enabling the segmentation model to reach strong performance with only a small fraction of labeled images.

\section{Conclusion}
\label{sec:concl}
In summary, we propose an appearance-prompted training-free pseudo-label generation and refinement framework for BUS segmentation. 
By leveraging simple and general appearance descriptions, the APPG module generates structurally consistent pseudo labels without additional training and initializes a static teacher to provide stable lesion priors. 
Combined with an EMA teacher and uncertainty-aware fusion and contrastive refinement, the framework effectively leverages unlabeled data and achieves performance comparable to fully supervised models under extremely limited annotations.

Beyond BUS, the proposed paradigm is readily extensible: for other modalities or diseases, only general appearance descriptions are required to obtain reliable pseudo supervision, enabling practical semi-supervised segmentation when annotations are scarce.

\bibliographystyle{IEEEtran}
\bibliography{IEEEabrv,reference}

\end{document}